\documentclass[manuscript]{acmart}
\setcopyright{none}
\AtBeginDocument{%
  }

\setcopyright{acmcopyright}
\copyrightyear{2018}
\acmYear{2018}
\acmDOI{XXXXXXX.XXXXXXX}

\acmConference[Conference acronym 'XX]{Make sure to enter the correct
  conference title from your rights confirmation emai}{June 03--05,
  2018}{Woodstock, NY}
\acmPrice{15.00}
\acmISBN{978-1-4503-XXXX-X/18/06}

\usepackage{caption}
\usepackage{subcaption}
\usepackage{adjustbox}
\usepackage{float}
\usepackage{booktabs} 
\usepackage{amsmath}
\usepackage{mathtools}
\usepackage{amsthm}

\usepackage[capitalize,noabbrev]{cleveref}
\usepackage{comment}
\usepackage[textsize=tiny]{todonotes}

\usepackage{algorithm}

\usepackage{xcolor}
\usepackage{algpseudocode}
	
\usepackage{soul}

\begin{document}

\newcommand\mycommfont[1]{\small\ttfamily\textcolor{blue}{#1}}
\title{Two-step counterfactual generation for OOD examples}


\author{Nawid Keshtmand}
\email{yl18410@bristol.ac.uk}
\affiliation{%
  \institution{University of Bristol}
  \streetaddress{Bristol BS8 1TL}
  \city{Bristol}
  \country{England}
}
\author{Raul Santos-Rodriguez}
\affiliation{%
  \institution{University of Bristol}
  \streetaddress{Bristol BS8 1TL}
  \city{Bristol}
  \country{England}
}
\email{enrsr@bristol.ac.uk}
\author{Jonathan Lawry}
\email{j.lawry@bristol.ac.uk}
\affiliation{%
  \institution{University of Bristol}
  \streetaddress{Bristol BS8 1TL}
  \city{Bristol}
  \country{England}
}


\renewcommand{\shortauthors}{Keshtmand et al.}


  \begin{abstract}
Two fundamental requirements for the deployment of machine learning models in safety-critical systems are to be able to detect out-of-distribution (OOD) data correctly and to be able to explain the prediction of the model. Although significant effort has gone into both OOD detection and explainable AI, there has been little work on explaining why a model predicts a certain data point is OOD. In this paper, we address this question by introducing the concept of an OOD counterfactual, which is a perturbed data point that iteratively moves between different OOD categories.
We propose a method for generating such counterfactuals, investigate its application on synthetic and benchmark data, and compare it to several benchmark methods using a range of metrics.
\end{abstract}


\maketitle

\section{Introduction}
In recent years, machine learning (ML) models have been increasingly deployed for prediction tasks. However, they still make erroneous predictions when exposed to inputs from an unfamiliar distribution. This poses a significant obstacle to the deployment of ML models in safety-critical applications such as healthcare and autonomous vehicles. Consequently, for applications in these domains, two fundamental requirements for the deployment of ML models are; 1) being able to identify data that is from a different distribution from the data on which the model was trained, which is referred to as out-of-distribution (OOD) detection, outlier detection, or anomaly detection \citep{yang2021generalized};
2) being able to explain the prediction of the model \citep{rudin2022interpretable}. There has been significant work on improving the accuracy of  OOD detectors although, there has not been much work on explaining why a data point is OOD \citep{pang2021toward}. As OOD detection algorithms are increasingly used in safety-critical domains, providing explanations for high-stakes decisions has become an ethical and regulatory requirement \citep{surden2021machine}. Therefore, it is important to develop methods that provide both accurate OOD scores and also  provide an explanation of why specific data points are detected as OOD. 

OOD detection can be considered a binary classification problem, where a data point can belong either to the in-distribution (ID) class or to the OOD class \cite{cui2022out}. Additionally, there are different versions of the OOD detection problem, which are referred to as near-OOD and far-OOD detection \citep{ren2021simple,winkens2020contrastive}.
 OOD data points that have neither non-discriminative (class-irrelevant) nor discriminative (class-relevant) features are referred to as far-OOD data and are therefore very dissimilar to the ID data. Whilst data points that contain similar non-discriminative features but do not contain discriminative features in common with the ID data are referred to as near-OOD data. As near OOD data points are more similar than far OOD data points to the ID data, near-OOD detection is generally considered to be more difficult than far-OOD detection \cite{winkens2020contrastive}.
An example of non-discriminative  and discriminative features is in images, where there are backgrounds (non-discriminative) plus objects (discriminative). OOD images may have both similar objects and a similar background to ID images. 

Results suggest that for near-OOD detection, standard methods of OOD detection such as the Mahalanobis Distance do not perform well \citep{lee2018simple}. Moreover, the proposed extension of the Mahalanobis Distance which takes into account class labels, and also the distribution fitted to the entire training data, does not work well in the far-OOD setting \citep{ren2021simple}. Huang et Al recently proposed a solution for OOD detection for both near and far-OOD detection which involves identifying two types of features in the latent space: discriminative and non-discriminative latent features. In their work, discriminative latent features are sufficient for distinguishing near-OOD data, while non-discriminative latent features are used for far-OOD detection \citep{huang2021decomposing}. Their approach involves decomposing the learned latent features into the two proposed types of latent features and then summing the Mahalanobis Distance estimated from each type of latent feature. This enabled their approach to be effective at both near and far-OOD detection. This showed the utility of non-discriminative and discriminative latent features for OOD detection.
We argue that the categories of near and far-OOD intuitively capture different extents to which OOD data differs from ID data and hence they can play a useful role in the development of explainable OOD detection. To explain the classification of a data point, it can be helpful to investigate the relevant decision boundary between classes. To get insight into why a data point is OOD, we believe it would be beneficial to look at investigating the decision boundary between a near-OOD, far-OOD, and an ID data point. An intuitive way to do this is to look for alternative inputs which are similar but are classified differently, These are referred to as counterfactuals \citep{molnar2020interpretable,poyiadzi2020face}. The usefulness of the counterfactual generated is related to how interpretable is this change in inputs. One way of perturbing the data is by changing several, if not all, latent features simultaneously. However, in this situation, ‘information overload’ is possible, i.e., too much evidence may cause confusion, reducing trust \citep{lahav2018interpretable}. We argue that changes in inputs of the following form are more interpretable:
\begin{itemize}
    \item Changes to individual latent features of the data.
    \item Changes to individual areas of interest (such as background or foreground object).
\end{itemize}
This is more likely to be true when the individual latent features are typically selected to be understandable. Therefore, we aim to limit the change in the counterfactual to only changing one area of interest at a time, where the areas we consider in this work are the discriminative and non-discriminative latent features. Thus in this paper, we study the problem of explaining OOD data points using the framework of counterfactual explanations. We introduce a definition of an OOD counterfactual as a perturbed data point that is generated by means of a two-step procedure that iteratively maps the point to move between far-OOD, near-OOD, and the ID categorization of data. We put forth a general framework for generating the counterfactuals and focus on a particular instantiation of the framework. The general framework involves changing the non-discriminative and discriminative latent features separately so that they are more similar to those of ID data. By having two distinct steps we can:
\begin{itemize}
    \item Generate counterfactuals across the different OOD categories.
    \item Examine how the (latent and/or input) features  of a data point change as it moves from far-OOD to near-OOD and finally to ID.
\end{itemize}

Understanding how the latent features are transformed by the two steps can help to differentiate between non-discriminative and discriminative latent features as well as indicate the extent to which certain latent features are responsible for the classification of a data point as OOD. 
The contributions of this paper are therefore as follows:
\begin{itemize}
    \item Introducing a new method for explainable OOD detection by generating counterfactual data points (Section \ref{Method}).
    \item Providing experimental results on tabular datasets showing that the proposed method generates counterfactuals that are more realistic and less likely to be identified as OOD compared to other baselines (Section \ref{Tabular_experiment}).
\end{itemize}

\section{Background}
\label{section:Background}
Throughout this paper, scalar mathematical quantities will be represented in lowercase, vectors will be in lowercase and bold, whilst matrix quantities will be capitalized and bold. The term 'class labels' will refer to the labels for the different classes of the ID dataset. The term 'OOD labels' will refer to whether the data point is ID or OOD. 



\subsection{Counterfactual explanations}
There are several different methods for explaining the predictions of an ML model. This includes local vs global approaches, anchors, SHAP, etc \citep{lundberg2017unified}. One popular approach is counterfactual explanations. Counterfactuals are used as post-hoc local explanations for individual decisions.
For an initial input vector $\boldsymbol{x}$ we look at asking the question why $\boldsymbol{x}$ is classified as class $c$ (or in our case why $\boldsymbol{x}$ is classified as being OOD). The framework of counterfactuals reformulates this "why" question so that it asks "what is the smallest change to $\boldsymbol{x}$ that would result in it belonging to a class that is different from $c$". A counterfactual explanation is an alternative input vector $\boldsymbol{x'}$, $\boldsymbol{x}' = \boldsymbol{x} + \boldsymbol{\delta}$ which involves $\boldsymbol{x}$ being perturbed in a particular way, where $\boldsymbol{\delta}$ is referred to as a perturbation.
This generally involves perturbing the input such that we can see the changes to the input features of $\boldsymbol{x}$ which would result in $\boldsymbol{x}$ crossing the decision boundary and being classified as a different class. This enables us to see which 
 input features are important for the classification of the data. Counterfactual explanations can be useful for debugging: They can be used to answer questions like ‘Why did the self-driving car misidentify the fire hydrant as a stop sign?’ and ’What would the image need to look like in order to be classified as a fire hydrant?’ \citep{goyal2019counterfactual,holtgen2021deduce}.

Depending on the model and the application, the requirements for the generated counterfactual may be different. One desired property of a counterfactual explanation is for the data to be realistic, i.e., a likely scenario for the user in question \citep{schut2021generating}.
Another commonly discussed requirement is sparsity: in general, the fewer features that are changed, the better. Sparse perturbations are usually more interpretable as the change in classification can be attributed to a smaller number of input features. This can make it easier to understand how the perturbation affects the model's prediction. If all the input features change a little, it can be harder to understand how each individual feature affects the model prediction. However, it is also possible to change a subset of the input features that are similar to one another, whilst keeping other input features the same. This could be viewed as imposing block sparsity.
\subsection{Counterfactual explanation approaches}
There are several different approaches to generating counterfactual explanations.
The Counterfactual Instances (CFI) approach \citep{wachter2017counterfactual} works as follows: $\boldsymbol{x}$ is the original instance vector whilst $\boldsymbol{x'}$ is the counterfactual. Let the model be given by $q$, and let $p_t$ be the target posterior probability of class $t$ (usually given a value of 1). The model's prediction on class $t$ is given by $q_{t}$. Let $\lambda$ be a hyperparameter. This method constructs counterfactual instances $\boldsymbol{x'}$ from an instance $\boldsymbol{x}$ by minimizing the following loss, $\mathcal{L}$, given in Eqn. \ref{eqn:CFI}:
\begin{equation}
\mathcal{L}(\boldsymbol{x'},\boldsymbol{x}) = (q_{t}(\boldsymbol{x'})-p_t)^2 + \lambda L_1(\boldsymbol{x'},\boldsymbol{x})    
\label{eqn:CFI}
\end{equation}
where $L_1$ is given by Eqn. \ref{eqn:L1 loss}: 
\begin{equation}
    L_{1}(\boldsymbol{x'},\boldsymbol{x}) = \left \| \boldsymbol{x'}-\boldsymbol{x} \right \|_1
    \label{eqn:L1 loss}
\end{equation}
The first term rewards the counterfactual for belonging to the desired class, and the use of the norm encourages sparse solutions. Various adaptations of this approach have also been proposed. Dhurandhar et Al train an auto-encoder to reconstruct instances of the training set \citep{dhurandhar2018explanations}. They then include the reconstruction error of the perturbed instance as an additional loss term in the objective function. As a result, the perturbed instance lies close to the training data manifold. The Counterfactuals Guided by Prototypes (Proto) approach by Van Looveren and Klaise includes an additional term in the loss function that optimizes for the distance between the latent features of a counterfactual instance $z'$ and a the latent features of 'prototypical' instance of the target class $z_{proto}$ \citep{looveren2021interpretable}. In doing this, it adds the additional requirement that interpretability also means that the generated counterfactual is close to a representative member of the target class.
\section{Method}
\label{Method}
\subsection{Overview} 
By framing OOD detection as a binary classification problem, where a data point, $\boldsymbol{x}$, can belong to the ID data class or the OOD data class, we can use the framework of counterfactuals explanations to investigate how the data point changes as it moves from OOD to ID. This can be used to explain why a data point is classified as OOD.
An example of this could be using a shape classifier for the purpose of classifying between circles and triangles. In the case where an ID dataset consists of red triangles and blue circles, a purple hexagon would be considered as a far-OOD data point. An initial counterfactual that would move the purple hexagon to a near-OOD data point, which is more similar to the ID data, would be to change the color to obtain a blue hexagon. Yet, despite having a similar color to the circles, the shape classifier would not be able to classify the hexagon as it is near-OOD and does not possess the same shape characteristics (sides, corners, or vertices) as a triangle or circle. Therefore, the next step of the OOD counterfactual generation process would be to change the blue hexagon to a blue circle so that it is an ID data point.
In the context of explaining why a data point is classified as OOD, we formulate counterfactual generation in terms of a perturbation $\boldsymbol{\delta}$ which satisfies Eqn. \ref{eqn:simple_ood_cf}:
\begin{equation}
    D_{ID}(f_z({\boldsymbol{x} + \boldsymbol{\delta}})) < D_{OOD}(f_{z}(\boldsymbol{x}+\boldsymbol{\delta}))
    \label{eqn:simple_ood_cf}
\end{equation}
where $D_{ID}$ and $D_{OOD}$ are metrics that represent how far the input is from being ID and OOD respectively, and $f_{z}$ is a latent feature extractor. But, as we do not have access to the OOD dataset during training time, we instead aim to find $\boldsymbol{\delta}$ which satisfies Eqn. \ref{eqn:approx_ood_cf}:
\begin{equation}
    D_{ID}(f_{z}(\boldsymbol{x} + \boldsymbol{\delta})) < D_{ID}(f_{z}(\boldsymbol{x}))
    \label{eqn:approx_ood_cf}
\end{equation}

The generation of a counterfactual for an OOD data point may involve perturbing the input so that the non-discriminative and/or discriminative latent features are more similar to those found in  ID data. If both non-discriminative and discriminative latent features are changed simultaneously this can make the overall transformation harder to interpret, as it would be difficult to tell which of the perturbed latent features are related to non-discriminative latent features and which are related to the discriminative latent features. Additionally, it would be difficult to tell how much each type of latent feature is responsible for the data point being classified as OOD.


Taking inspiration from Huang et Al, we propose that an OOD counterfactual should be able to provide explanations for both near and far-OOD data points \citep{huang2021decomposing}. Therefore, it is desired that the OOD data point is perturbed such that both non-discriminative and discriminative latent features of the counterfactual become more similar to that of the ID data but the non-discriminative and discriminative latent features are perturbed in separate steps.
In this case, we formalize an OOD counterfactual as a perturbed data point that is generated in a two-step procedure that involves changing the non-discriminative followed by the discriminative latent features, or discriminative followed by non-discriminative latent features.
More formally, a latent vector $\boldsymbol{z}$ is separated into a non-discriminative part, $\boldsymbol{z_{n}}$, and discriminative part $\boldsymbol{z_{d}}$ i.e., $\boldsymbol{z} = (\boldsymbol{z_n}; \boldsymbol{z_d})$. Then we can satisfy Eqn. \ref{eqn:approx_ood_cf} by breaking it down into two steps given by Eqn. \ref{eqn:cf_non_dis} and \ref{eqn:cf_dis}:
\begin{equation}
    D_{ID}(f_{z}(\boldsymbol{x} + \boldsymbol{\delta_{n}})) < D_{ID}(f_{z}(\boldsymbol{x}))
    \label{eqn:cf_non_dis}
\end{equation}

\begin{equation}
    D_{ID}(f_{z}(\boldsymbol{x} + \boldsymbol{\delta_{d}})) < D_{ID}(f_{z}(\boldsymbol{x}))
    \label{eqn:cf_dis}
\end{equation}
where $\boldsymbol{\delta_{n}}$ and $\boldsymbol{\delta_{d}}$ correspond to the perturbation to the non-discriminative and discriminative feature partitions respectively. Eqn. \ref{eqn:approx_ood_cf} is satisfied by having $\boldsymbol{\delta} = \boldsymbol{\delta_n} + \boldsymbol{\delta_d}$. Performing the counterfactual generation process in this way, explicitly ensures that the generated counterfactuals are similar to the ID data in both the non-discriminative and discriminative latent features. In addition, the counterfactuals can be made to change data points to different degrees of OOD, such as going from far-OOD to ID, near-OOD to ID or even far-OOD to near-OOD. This can improve interpretability since a user can examine the change in a certain step to see whether there is a large or small change to a particular latent feature partition to become more ID. A large change would indicate that the data point was OOD due to that subset of latent features being far away from the ID data. If there was only a single set of latent features, then it may not be possible to find out whether it is the non-discriminative and/or discriminative latent features causing the data point being classified as OOD.


In order to generate the OOD counterfactual, there are three general parts required:
\begin{itemize}
    \item Obtain latent features of the OOD data, $\boldsymbol{z}$.
    \item Decompose $\boldsymbol{z}$ into two partitions, $\boldsymbol{z_{d}}$ and $\boldsymbol{z_{n}}$, using a classifier.
    \item Perturb the OOD data, $\boldsymbol{x}$, in a way that makes each latent partition more similar to the latent feature partitions of the ID data.
\end{itemize}
Therefore a user can generate an OOD counterfactual using any approach which satisfies these three steps.
\subsection{Obtaining and decomposing features}
Whilst Huang et Al obtain latent features of the data using a pre-trained neural network and then decompose the latent features using an independent cross-entropy (ICE) classification loss, we take an alternative approach \citep{huang2021decomposing}. We decided to use one of the baselines from the work of Huang et Al, which was to use Principal Components Analysis (PCA) to obtain the latent features, $\boldsymbol{z}$. We used PCA as we wanted to obtain latent features which maximize the variance in the data. We additionally separate the principal components into a non-discriminative partition, $\boldsymbol{z_{n}}$, and a discriminative partition $\boldsymbol{z_{d}}$.

To separate the data into partitions, we examine the conditional entropy of a subset of the principal components $\boldsymbol{Z_{subset}}$ with the class labels $\boldsymbol{Y}$, $H[\boldsymbol{Y}|\boldsymbol{Z_{subset}}]$, using Eqn. \ref{eqn:conditional_entropy} \citep{thomas2006elements}:
\begin{equation}
H[\boldsymbol{Y}|\boldsymbol{Z_{subset}}] = \mathbb{E}_{\boldsymbol{z_{subset}}\sim \boldsymbol{Z_{subset}}} [h(P(\boldsymbol{y=1}|\boldsymbol{z_{subset}})]
\label{eqn:conditional_entropy}
\end{equation}
where $\boldsymbol{y}$ corresponds to the label for a single data point and $\boldsymbol{z_{subset}}$ corresponds to the latent feature (principal component) subset for a single data point. Also, $h(p)= - p \log p - (1 - p)\log(1 - p)$ is the entropy of a Bernoulli distribution. A variety of different classifiers could be used to obtain the Bernoulli distribution such as a logistic regression classifier. In our work, we used a Quadratic Discriminant Analysis (QDA) classifier \citep{hastie2009elements}. This involved fitting a Gaussian mixture model (GMM) to the latent features of each partition. Then, the minimal subset $\boldsymbol{z_{d}}$ is obtained using Eqn. \ref{eqn:scaled_subset}:
\begin{equation}
    \boldsymbol{z_{d}} = \min_{\boldsymbol{Z_{subset}}}
    \mathcal{L}_{Partition}(\boldsymbol{Z_{subset}},\boldsymbol{\hat{Y}_{subset}}, \boldsymbol{\hat{Y}_{subset}^{\complement}})
    \label{eqn:scaled_subset}
\end{equation}
where $\mathcal{L}_{Partition}$ is given by Eqn: \ref{Eqn:Partition_loss} 
\begin{equation}
     \mathcal{L}_{Partition}(\boldsymbol{Z_{subset}},\boldsymbol{\hat{Y}_{subset}}, \boldsymbol{\hat{Y}_{subset}^{\complement}}) = \\ H[\boldsymbol{\hat{Y}_{subset}}|\boldsymbol{Z_{subset}}]-H[\boldsymbol{\hat{Y}_{subset}^{\complement}}|\boldsymbol{Z_{subset}^{\complement}}]
     \label{Eqn:Partition_loss}
\end{equation}

where $\boldsymbol{Z}_{subset}^{\complement}$ includes all the latent features not in $\boldsymbol{Z_{subset}}$, $\boldsymbol{\hat{Y}_{subset}}$ and $\boldsymbol{\hat{Y}_{subset}^{\complement}}$ correspond to predictions of a classifier trained on 
$\boldsymbol{Z_{subset}}$ and a classifier trained on $\boldsymbol{Z_{subset}^{\complement}}$ respectively. 

After finding $\boldsymbol{z_d}$, the non-discriminative partition $\boldsymbol{z_{n}}$ is then given by the remaining latent features which are not in the subset $\boldsymbol{z_{d}}$. The motivation of this loss is that we want to find a $\boldsymbol{Z_{subset}}$ which has a low conditional entropy with $\boldsymbol{Y}$, as this would suggest that the latent features in $\boldsymbol{Z_{subset}}$ are strongly correlated with $\boldsymbol{Y}$ and would correspond to discriminative latent features. Furthermore, we want to find a $\boldsymbol{Z_{subset}^{\complement}}$ with high conditional entropy as this would mean that the latent features in $\boldsymbol{Z_{subset}^{\complement}}$ are not strongly correlated with $\boldsymbol{Y}$ and therefore likely to correspond to non-discriminative latent features.
But, an issue with the loss is that it will tend to separate most of the latent features into $\boldsymbol{z_d}$ as the entropy of $\boldsymbol{z_d}$ will tend to decrease as more latent features are added into $\boldsymbol{z_d}$ and the entropy of $\boldsymbol{z_{n}}$ will increase as fewer latent features are in $\boldsymbol{z_{n}}$.
Moreover, it seems that adding additional latent features into $\boldsymbol{Z_{subset}}$ does not lead to $\mathcal{L}_{Partition}$ decreasing by a significant amount. To take into account the bias that more latent features are placed in $\boldsymbol{Z_{subset}}$ generally lowers the entropy, we also add a threshold such that if the difference between two subsets with different cardinality is small, then we choose $\boldsymbol{z_d}$ to be the subset which has the smaller cardinality. We did this as we believe that we should maximize the number of latent features that are present in the $\boldsymbol{z_n}$ as in a realistic classification task, a large number of the features will tend to be unrelated to a classification task. The procedure for generating to separate the features is described further in Appendix .\ref{Appendix: Feature partitioning}.

\subsection{Perturbing OOD data}
After obtaining the latent features and decomposing them into $\boldsymbol{z_n}$ and $\boldsymbol{z_d}$, counterfactuals can be generated by perturbing OOD data such that the $\boldsymbol{z_{n}}$ and $\boldsymbol{z_{d}}$ of the counterfactuals are similar to values of  $\boldsymbol{z_n}$ and $\boldsymbol{z_d}$ of the ID data. As Gaussian-based approaches such as the Mahalanobis Distance and maximising the likelihood of a data point with respect to a gaussian performs well at OOD detection,  we decided that a counterfactual with a low negative log-likelihood (NLL) under a gaussian mixture model (GMM) obtained from the ID data would mean that the counterfactual is similar to the ID data points. Therefore we believe that minimizing the NLL of data point under a GMM would be an effective loss function to optimize for. Furthermore, as we generated the partitions of the data by using QDA, which is a Gaussian-based classification approach, it would also be beneficial to optimize for a Gaussian-based OOD detection approach. By doing this, there is a better match between the way the partitions were generated and the generation of the counterfactuals.
This involved modeling the data distribution of the PCA feature space by a separate Gaussian distribution for $\boldsymbol{z_n}$ and $\boldsymbol{z_d}$. 
Counterfactuals are perturbed to minimize the NLL under the GMM defined by  non-discriminative partition in the first step of the process and then perturbed to minimize the NLL under the GMM defined by the discriminative partition in the second step of the process. Although, it is possible to optimize for the discriminative partition first and then the non-discriminative partition. The loss for the non-discriminative, $\mathcal{L}_{CF,n}$ and discriminative partitions $\mathcal{L}_{CF,d}$ are given by \ref{eqn:counterfactual_loss_non_discriminative} and \ref{eqn:counterfactual_loss_discriminative} respectively: 
\begin{equation}
    \mathcal{L}_{CF,n} = -\log p_{n}(f_z(\boldsymbol{x})) 
    \label{eqn:counterfactual_loss_non_discriminative}
\end{equation}

\begin{equation}
    \mathcal{L}_{CF,d} = -\log p_{d}(f_z(\boldsymbol{x})) 
    \label{eqn:counterfactual_loss_discriminative}
\end{equation}
where $p_n$ and $p_d$  is the density under the non-discriminative and discriminative GMM respectively.  Also, $f_z$ is the feature extractor which in this case is the PCA algorithm.
We refer to the loss for the non-discriminative and
discriminative partitions as the Non-discriminative (Non-dis) and Discriminative (Dis) loss respectively.
The process of generating the counterfactuals can be seen in Algorithms \ref{alg:counterfactual_OOD_natural_language} and \ref{alg:counterfactual_OOD_pseudocode} in Appendix \ref{Appendix: Counterfactual_generation}.

\section{2D Toy Experiment}
\label{Toy_experiment}
\begin{figure}[h!]
    \centering
    \includegraphics[width=0.4\textwidth]{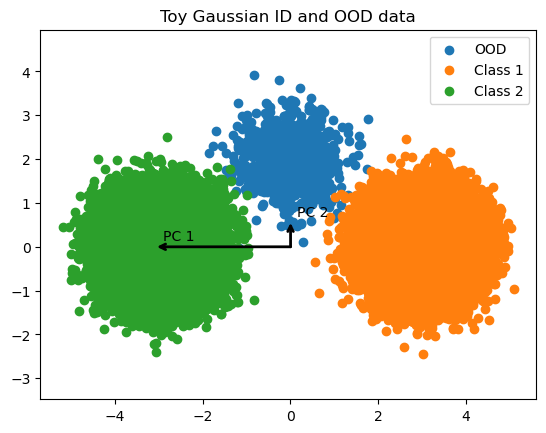}
    \caption{Toy dataset showing the ID data, OOD data, and Principal components of ID data.}
    \label{fig:Principal_components}
\end{figure}

We consider a low dimensional (2D) simulation where we generate two Gaussian distributions with a mean (3,0) and (-3,0) with a tied covariance matrix $\boldsymbol{\Sigma} = 0.5\boldsymbol{I_{2}}$. There is also an additional OOD distribution which has a mean (0,2) and covariance $\boldsymbol{\Sigma} = 0.3\boldsymbol{I_{2}}$. PCA with two principal components is then used to extract the latent features which capture the most variance in the ID data. The setup of the toy dataset and the principal components can be seen in Fig. \ref{fig:Principal_components}. Each latent feature is evaluated for how effective it is at capturing the discriminative information in the ID dataset using the conditional entropy $H\left [\boldsymbol{Y}| \boldsymbol{X} \right ]$, where $\boldsymbol{pc1}$ is shown to have a  $H\left [\boldsymbol{Y}| \boldsymbol{pc1} \right ]$ =$1.92e^{-3}$ whilst $\boldsymbol{pc2}$ has a $H\left [\boldsymbol{Y}| \boldsymbol{pc2} \right ] =1.00$.

\begin{figure}[h!]
     \centering
     \begin{subfigure}[h!]{0.49\textwidth}
         \centering\includegraphics[width=\textwidth]{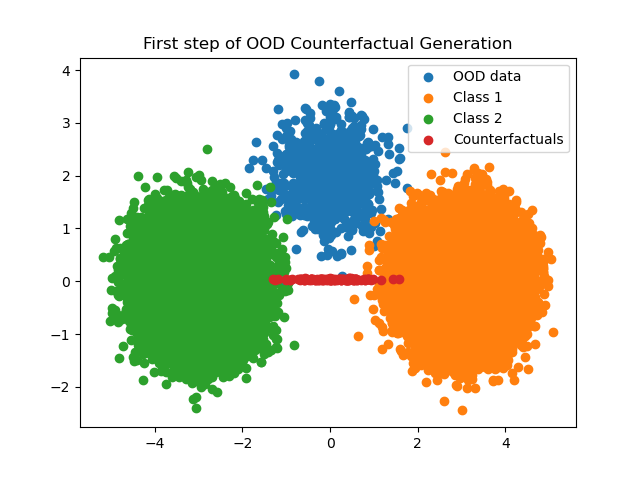}
         \caption{Optimizing for $\boldsymbol{z_n}$.}
         \label{fig:zn_zd_first_step}
     \end{subfigure}
     \hfill
     \begin{subfigure}[h!]{0.49\textwidth}
         \centering
         \includegraphics[width=\textwidth]{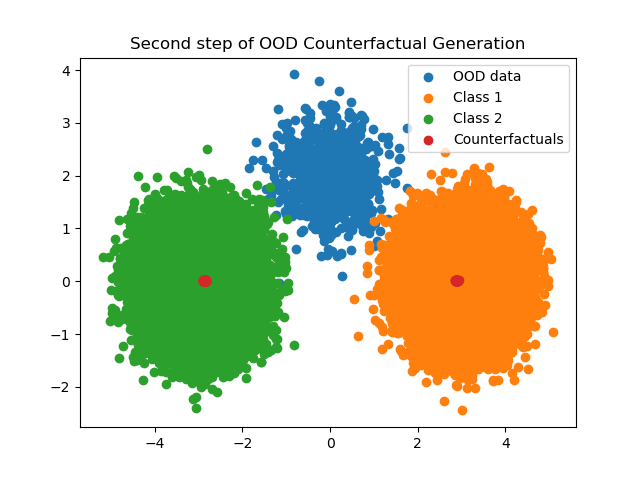}         
         \caption{Optimizing for $\boldsymbol{z_d}$.}
         \label{fig:zn_zd_second_step}
     \end{subfigure}
\caption{OOD counterfactual generation when optimizing for $\boldsymbol{z_n}$ then $\boldsymbol{z_d}$.}
\label{fig:nuisance-class}
\end{figure}

\begin{figure}[h!]
     \centering
     \begin{subfigure}[h!]{0.49\textwidth}
         \centering\includegraphics[width=\textwidth]{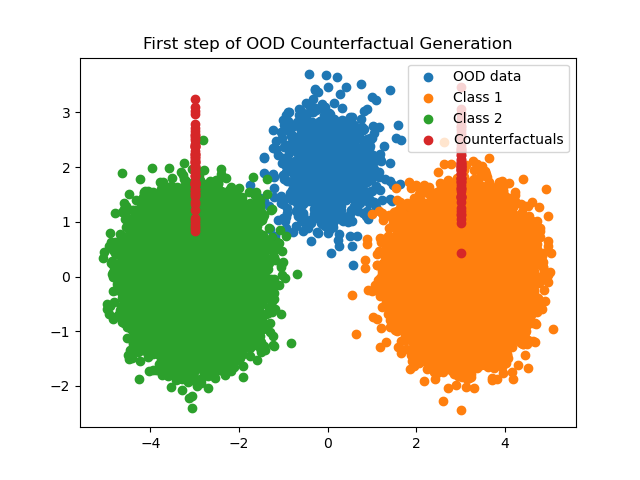}
         \caption{Optimizing for $\boldsymbol{z_d}$.}
         \label{fig:zd_zn_first_step}
     \end{subfigure}
     \hfill
     \begin{subfigure}[h!]{0.49\textwidth}
         \centering
         \includegraphics[width=\textwidth]{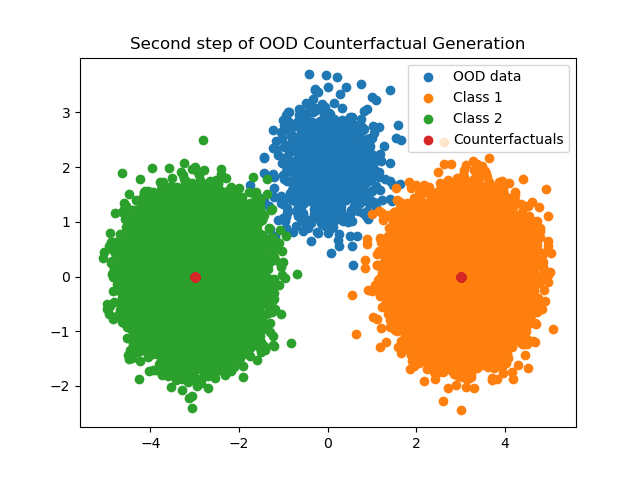}         
         \caption{Optimizing for $\boldsymbol{z_n}$.}
         \label{fig:zd_zn_second_step}
     \end{subfigure}
\caption{OOD counterfactual generation when optimizing for $\boldsymbol{z_d}$ then $\boldsymbol{z_n}$.}
\label{fig:class-nuisance}
\end{figure}

After obtaining the principal components, an OOD counterfactual can be produced using Eqn. \ref{eqn:counterfactual_loss_non_discriminative} and \ref{eqn:counterfactual_loss_discriminative}. The counterfactual produced depends on whether $\boldsymbol{z_n}$ or $\boldsymbol{z_d}$ is optimized first in the two-step optimization process. Fig. \ref{fig:nuisance-class} shows the generation of the counterfactuals by first optimizing for $\boldsymbol{z_n}$ then $\boldsymbol{z_d}$, whilst Fig. \ref{fig:class-nuisance} shows the generation of the counterfactuals by first optimizing for $\boldsymbol{z_d}$ then $\boldsymbol{z_n}$. In Fig. \ref{fig:nuisance-class}, it can clearly be seen that the counterfactuals initially move in the vertical direction, which does not aid in the classification of the data points. In the second step, the data points are moved in the horizontal direction which enables them to be easily classified. Having both steps present clearly shows that non-discriminative and discriminative latent features are important for deciding whether a data point is OOD.

\begin{figure}[h!]
     \centering
     \begin{subfigure}[h!]{0.49\textwidth}
\centering\includegraphics[width=\textwidth]{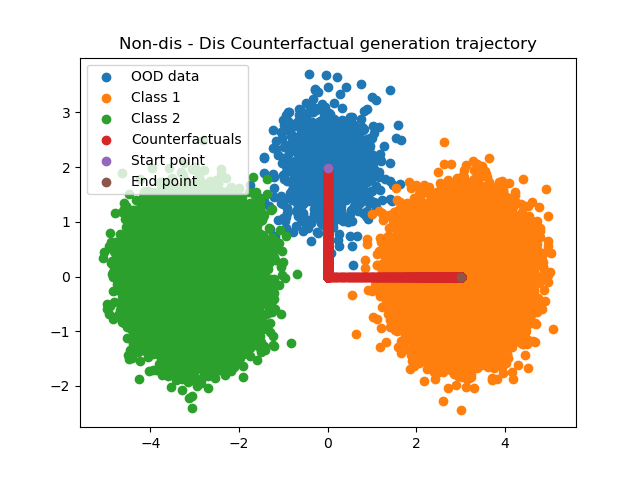}
         \caption{Optimizing for $\boldsymbol{z_n}$ and then $\boldsymbol{z_d}$.}
         \label{fig:non-dis dis trajectory}
     \end{subfigure}
     \hfill
     \begin{subfigure}[h!]{0.49\textwidth}
         \centering         \includegraphics[width=\textwidth]{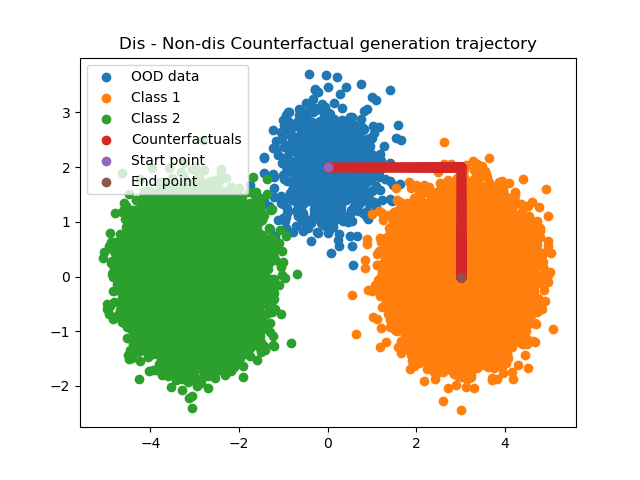}  
         \caption{Optimizing for $\boldsymbol{z_d}$ and then $\boldsymbol{z_n}$.}
         \label{fig:dis non-dis trajectory}
     \end{subfigure}
\caption{Optimization trajectory of a single OOD data point}
\label{fig:cf_trajectory}
\end{figure}

For additional clarity of how the counterfactual changes during the optimization process, we have also shown the optimization trajectory taken by a single data point in Fig . \ref{fig:cf_trajectory}. In Fig. \ref{fig:non-dis dis trajectory}, an initial point at position (0,2) moves towards the centroid of Class 1 by first moving along the vertical (non-discriminative) direction then along the horizontal (discriminative) direction. Whilst in Fig \ref{fig:dis non-dis trajectory}, the data point moves along the horizontal (discriminative) and then vertical (non-discriminative) directions.  


To get an indication of whether optimizing for Eqn. \ref{eqn:counterfactual_loss_non_discriminative} and \ref{eqn:counterfactual_loss_discriminative} would
generate counterfactuals that are indistinguishable from the ID data, we examined the OOD detection performance  when using  $\mathcal{L}_{Total} = \mathcal{L}_{CF,n} + \mathcal{L}_{CF,d}$ as a metric for distinguishing between OOD and ID data. This was done by calculating OOD scores using $\mathcal{L}_{Total}$ and then calculating the AUROC from the OOD scores. 
The AUROC measures the Area Under the Receiver Operating Characteristic curve. The Receiver Operating Characteristic (ROC) curve plots the relationship between the true positive rate (TPR) and false positive rate (FPR). The area under the ROC curve can be interpreted as the probability that a positive example (ID) will have a higher detection score than a negative example (OOD). The OOD detection performance was compared with the Mahalanobis Distance and the Marginal Mahalanobis Distance. The Marginal Mahalanobis Distance is a variant of the Mahalanobis Distance which fits a single Gaussian to the entire training dataset. From the results in Table. \ref{table:Toy_OOD}, it can be seen that  $\mathcal{L}_{Total}$, was the most effective metric at detecting OOD data, and therefore optimizing this objective should lead to the generation of realistic OOD counterfactuals.


\begin{table}[h!]
\begin{tabular}{|l|l|}
\hline
OOD approach                               & AUROC \\ \hline
Mahalanobis Distance & 0.908 \\ \hline
Marginal Mahalanobis Distance              & 0.908 \\ \hline
Custom metric                              & 0.999 \\ \hline
\end{tabular}
\caption{Toy OOD detection results}
\label{table:Toy_OOD}
\end{table}

\section{Tabular data experiment}
\label{Tabular_experiment}

\subsection{Evaluation Metrics}
To evaluate the counterfactual generation algorithms, we evaluate how realistic the counterfactuals are and how minimal they have changed from the original data points.
To measure the minimality, we calculated the L1 distance between data points, $\boldsymbol{x}$, and counterfactuals $\boldsymbol{x'}$ given by Eqn. \ref{eqn:L1 loss} as shown in Section \ref{section:Background}. 
Realism is more difficult to measure because it is poorly defined and in the literature, there are many different definitions used. In this work, we define realism as a counterfactual that is similar to the ID data and therefore similar in the non-discriminative and discriminative latent features of the ID data. Thus, we desire the counterfactuals to have a low NLL for the non-discriminative and the discriminative latent features as given by Eqn. \ref{eqn:counterfactual_loss_non_discriminative} and \ref{eqn:counterfactual_loss_discriminative}. To measure this, we calculate the AUROC between the ID data (positive examples) and the counterfactual generated from the OOD data (negative examples). In this case, counterfactuals which lead to lower AUROC values are less distinguishable from the ID data and thus more realistic counterfactuals. Therefore, in contrast to the previous use of AUROC in the OOD detection literature, in this work, approaches that have lower AUROC are better for this metric. Additionally, we examined how well the different approaches optimized the Non-dis and Dis loss given by Eqn. \ref{eqn:counterfactual_loss_non_discriminative} and \ref{eqn:counterfactual_loss_discriminative}.
For each of the metrics shown, the lower the value of the metric, the better the approach of interest is in terms of that metric. Unless otherwise stated, metrics reported in this work are obtained based on 5 repeat readings.

\subsubsection{Datasets}
We perform our analysis on five tabular datasets: Thyroid, Heart, Pima Diabetes, Wine, and Cardiotocography dataset. We use tabular datasets as this type of data is frequently used in the interpretability literature. To be able to use the approach for OOD counterfactual generation, it was important that the datasets were preprocessed in some manner to have different classes as well as outliers that can be used for counterfactual generation. Additional information on the datasets and how they were preprocessed can be seen in Appendix \ref{Appendix: Datasets}.
\subsection{Results and Discussion}

\begin{minipage}[c]{0.5\textwidth}
\centering
\begin{tabular}{|c|c|c|c|c|}
\hline
 Approach & Non-dis & Dis & L1 & AUROC \\
\hline
 OOD CF &     1.444 & -2.230 &  11.834 &        0.860  \\
Proto  &     1.646 &  7.798 &  10.620 &        0.994  \\
CFI    &     1.662 &  8.568 &   9.598 &        1.000  \\
CEM    &     1.726 &  6.322 &   8.756 &        1.000  \\
\hline
\end{tabular}
\captionof{table}{Base metrics-Wine}
\label{table:Base metrics-Wine}
\end{minipage}
\begin{minipage}[c]{0.5\textwidth}
\centering
\begin{tabular}{|c|c|c|c|c|}
\hline
 Approach & Non-dis & Dis & L1 & AUROC \\
\hline
 OOD CF &    -0.260 &  1.258 &  3.888 &        0.052  \\
Proto  &    -0.070 &  1.928 &  5.342 &        0.364  \\
CFI    &     2.624 &  2.290 &  5.974 &        0.586  \\
CEM    &     1.314 &  2.170 &  5.580 &        0.556  \\
\hline
\end{tabular}
\captionof{table}{Base metrics-Diabetes}
\label{table:Base metrics-Diabetes}
\end{minipage}
\begin{table}[h!]
\centering
\begin{tabular}{|c|c|c|c|c|}
\toprule
Approach &  Non-dis &  Dis &      L1 &  AUROC  \\
\midrule
OOD CF &     0.752 & -2.718 &   8.874 &        0.382  \\
Proto  &     0.734 &  0.370 &  10.484 &        0.384  \\
CFI    &     0.994 &  2.238 &  10.930 &        0.580  \\
CEM    &     0.968 &  1.948 &  10.654 &        0.606  \\
\bottomrule
\end{tabular}
\caption{Base metrics-Heart}
\label{table:Base metrics-HeartOOD}
\end{table}

\begin{minipage}[c]{0.5\textwidth}
\centering
\begin{tabular}{|c|c|c|c|c|}
\hline
 Approach & Non-dis & Dis & L1 & AUROC \\
\hline
 OOD CF &    -0.538 &  1.096 &  3.552 &        0.312  \\
Proto  &     0.204 &  1.284 &  9.726 &        0.686  \\
CFI    &     1.628 &  1.876 &  9.922 &        0.800  \\
CEM    &     0.972 &  2.062 &  9.984 &        0.896  \\
\hline
\end{tabular}
\captionof{table}{Base metrics-Thyroid}
\label{table:Base metrics-Thyroid}
\end{minipage}
\begin{minipage}[c]{0.5\textwidth}
\centering
\begin{tabular}{|c|c|c|c|c|}
\hline
 Approach & Non-dis & Dis & L1 & AUROC \\
\hline
OOD CF &     1.532 & -5.118 &  10.926 &        0.918  \\
Proto  &     1.918 & -0.078 &  21.124 &        0.924  \\
CFI    &     2.054 &  4.894 &  21.956 &        0.898  \\
CEM    &     1.986 &  1.886 &  22.398 &        0.946  \\
\hline
\end{tabular}
\captionof{table}{Base metrics-Cardio}
\label{table:Base metrics-Cardio}
\end{minipage}

We compared the OOD counterfactual (CF) approach with state-of-the-art baselines which are frequently used in the counterfactuals literature. The three baselines are:
\begin{itemize}
\item Counterfactual Instances (CFI) \citep{wachter2017counterfactual}
\item Counterfactuals Guided by Prototypes (Proto) \citep{looveren2021interpretable}
\item Contrastive Explanation Method (CEM) \citep{dhurandhar2018explanations}
\end{itemize}
To generate counterfactuals with the baselines, we use the authors’ implementation in the Alibi package \cite{JMLR:v22:21-0017}. All the baselines involve training a neural network classifier which contains 2 hidden layers with dimensionality equal to the input dimension and a number of classes equal to the number of ID classes of the data. The classifiers were trained for 500 epochs, using the SGD optimizer with a batch size of 128 and a learning rate of 0.01.

From Tables \ref{table:Base metrics-Wine} - \ref{table:Base metrics-Cardio}, it can be seen that the Dis loss is consistently lower for the OOD CF approach compared to the baselines. Moreover, the Non-dis loss is lower for OOD CF for all datasets except the Heart Dataset. This shows that the approach effectively optimizes both of the metrics of interest. Moreover, despite not having any explicit terms in the optimization to minimize the $L_{1}$ distance, it can be seen that OOD CF is able to outperform all the baselines in minimizing the $L_{1}$ distance for all datasets except for the Wine dataset. This could suggest that separating the counterfactual generation into two distinct steps, makes it so that the counterfactuals produced have a lower $L_{1}$ distance to the original instances. In addition, it can be seen that the AUROC when using the OOD CF approach is significantly better for the Wine, Diabetes as well as Thyroid datasets. Whilst for the larger dimensionality datasets (Cardio, and Heart), the AUROC value is similar to the best-performing baseline. This suggests that having lower values for the metrics of interest leads to more realistic counterfactuals which also have lower $L_{1}$ distances compared to the baselines. This is despite the baselines explicitly regularising to minimize the $L_{1}$ distance whilst OOD CF does not. The effectiveness of a counterfactual generation approach that produces realistic ID data points is likely related to the degree to which the density of the ID data points is taken into account. The approach which leads to the least realistic counterfactuals is the CFI approach, which does not take into account the density of data points but instead focuses on generating counterfactuals that are far across the decision boundary. The best-performing baseline is the prototype approach which generates counterfactuals that are near to the prototype of the class. This can be seen as an approach that implicitly maximizes the probability density (likelihood) of a counterfactual. Our approach, which generally outperforms the baselines, can be seen to explicitly maximize the probability density (likelihood) of the counterfactual which consequently leads to the most realistic counterfactuals.


\subsection{Ablation - Single Gaussian Counterfactual}
\begin{minipage}[c]{0.5\textwidth}
\centering
\begin{tabular}{|c|c|c|c|c|}
\hline
 Approach & Non-dis & Dis & L1 & AUROC \\
\hline
OOD CF                 &     1.444 &  -2.230 &  11.834 &        0.860  \\
OOD SG &     1.558 &  -1.602 &  13.218 &        0.916  \\
\hline
\end{tabular}
\captionof{table}{Single Gaussian-Wine}
\label{table:Single Gaussian-Wine}
\end{minipage}
\begin{minipage}[c]{0.5\textwidth}
\centering
\begin{tabular}{|c|c|c|c|c|}
\hline
 Approach & Non-dis & Dis & L1 & AUROC \\
\hline
OOD CF                 &    -0.260 &  1.258 &  3.888 &        0.052  \\
OOD SG &    -0.254 &  2.334 &  7.472 &        0.190  \\
\hline
\end{tabular}
\captionof{table}{Single Gaussian-Diabetes}
\label{table:Single Gaussian-Diabetes}
\end{minipage}

\begin{table}[h!]
\centering
\begin{tabular}{|c|c|c|c|c|}
\hline
Approach &  Non-dis &  Dis &      L1 &  AUROC  \\
\hline
OOD CF                 &     0.752 & -2.718 &   8.874 &        0.382  \\
OOD SG &     0.682 &  1.446 &  11.868 &        0.888  \\
\hline
\end{tabular}
\caption{Single Gaussian-Heart}
\label{table:Ablation-single gaussian-HeartOOD}
\end{table}

\begin{minipage}[c]{0.5\textwidth}
\centering
\begin{tabular}{|c|c|c|c|c|}
\hline
 Approach & Non-dis & Dis & L1 & AUROC \\
\hline
OOD CF                 &    -0.538 &  1.096 &   3.552 &        0.312  \\
OOD SG &    -0.450 &  1.094 &  10.586 &        0.232  \\
\hline
\end{tabular}
\captionof{table}{Single Gaussian-Thyroid}
\label{table:Single Gaussian-Thyroid}
\end{minipage}
\begin{minipage}[c]{0.5\textwidth}
\centering
\begin{tabular}{|c|c|c|c|c|}
\hline
 Approach & Non-dis & Dis & L1 & AUROC \\
\hline
OOD CF                 &     1.532 & -5.118 &  10.926 &        0.918  \\
OOD SG &     1.518 & -4.428 &  19.538 &        0.978  \\
\hline
\end{tabular}
\captionof{table}{Single Gaussian-Cardio}
\label{table:Single Gaussian-Cardio}
\end{minipage}

To examine the effectiveness of separating the latent features into two partitions, we performed an ablation that examined generating counterfactuals that are optimized to have a high likelihood under a single GMM. This approach is referred to as OOD Single Gaussian (SG). From Table \ref{table:Single Gaussian-Wine} - \ref{table:Single Gaussian-Cardio}, it can be seen that the results of the $L_{1}$ distance are consistently lower for the case where there are two separate partitions of the latent features. Moreover, the AUROC values are lower for OOD CF for all datasets except the Thyroid dataset. This could be due to the Thyroid dataset having lower values for the Non-dis and Dis loss for the OOD SG. This further suggests that low values of Non-dis and Dis loss correlate well with the OOD detection performance. Hence, it can be seen that by separating the latent features into two distinct partitions, we are able to obtain counterfactuals that have lower $L_{1}$ distances whilst also being more realistic compared to using a single partition.

\subsection{Ablation - Only Single Step Counterfactual}
\begin{minipage}[c]{0.5\textwidth}
\centering
\begin{tabular}{|c|c|c|c|c|}
\hline
 Approach & Non-dis & Dis & L1 & AUROC \\
\hline
OOD CF                   &     1.444 & -2.230 &  11.834 &        0.860 \\
OOD SN       &     1.444 &  5.658 &   2.712 &        0.998 \\
OOD SD &     1.754 & -2.230 &  11.188 &        0.920 \\
\hline
\end{tabular}
\captionof{table}{Single Step-Wine}
\label{table:Single Step-Wine}
\end{minipage}
\begin{minipage}[c]{0.5\textwidth}
\centering
\begin{tabular}{|c|c|c|c|c|}
\hline
 Approach & Non-dis & Dis & L1 & AUROC \\
\hline
OOD CF                   &    -0.260 &  1.258 &  3.888 &        0.052 \\
OOD SN       &    -0.260 &  2.104 &  1.272 &        0.344 \\
OOD SD &     0.822 &  1.258 &  3.432 &        0.162 \\
\hline
\end{tabular}
\captionof{table}{Single Step-Diabetes}
\label{table:Single Step-Diabetes}
\end{minipage}
\begin{table}[h!]
\centering
\begin{tabular}{|c|c|c|c|c|}
\hline
Approach &  Non-dis &  Dis &     L1 &  AUROC \\
\hline
OOD CF                   &     0.752 & -2.718 &  8.874 &        0.382 \\
OOD SN       &     0.752 &  1.420 &  1.880 &        0.404 \\
OOD SD &     1.068 & -2.718 &  8.560 &        0.350 \\
\hline
\end{tabular}

\caption{Single Step-Heart}
\label{table:Ablation-single step-HeartOOD}
\end{table}

\begin{minipage}[c]{0.5\textwidth}
\centering
\begin{tabular}{|c|c|c|c|c|}
\hline
 Approach & Non-dis & Dis & L1 & AUROC \\
\hline
OOD CF                   &    -0.538 &  1.096 &  3.552 &        0.312 \\
OOD SN       &    -0.538 &  2.008 &  1.668 &        0.600 \\
OOD SD &     0.488 &  1.096 &  2.706 &        0.640 \\
\hline
\end{tabular}
\captionof{table}{Single Step-Thyroid}
\label{table:Single Step-Thyroid}
\end{minipage}
\begin{minipage}[c]{0.5\textwidth}
\centering
\begin{tabular}{|c|c|c|c|c|}
\hline
 Approach & Non-dis & Dis & L1 & AUROC \\
\hline
OOD CF                   &     1.532 & -5.118 &  10.926 &        0.918 \\
OOD SN       &     1.532 & -1.188 &   3.360 &        0.868 \\
OOD SD &     1.982 & -5.118 &  10.176 &        0.896\\
\hline
\end{tabular}
\captionof{table}{Single Step-Cardio}
\label{table:Single Step-Cardio}
\end{minipage}

To examine the effectiveness of optimizing for the non-discriminative and discriminative partitions, we performed an ablation that examined generating counterfactuals that are optimized for only one of the two partitions. The approaches are OOD Single Non-dis (SN) and OOD Single Dis (SD) for the case where only the non-discriminative and discriminative partitions are optimized respectively. By comparing the situation between optimizing both partitions and optimizing for a single partition, it can be seen that optimizing for the discriminative partition leads to a much larger change in the $L_{1}$ distance compared to optimizing for the non-discriminative partition.
Also, the AUROC values for optimizing both partitions compared to a single partition are much better for the lower dimensional datasets (Wine, Diabetes, and Thyroid) as seen from Tables \ref{table:Single Step-Wine}, \ref{table:Single Step-Diabetes} and \ref{table:Single Step-Thyroid}, but are similar for the case of the higher dimensional datasets (Cardio and Heart) as seen in Table \ref{table:Ablation-single step-HeartOOD} and \ref{table:Single Step-Cardio}. 
Thus, the benefit of using two steps is more apparent in the case of lower-dimensional datasets.
\subsection{Limitations}
Our approach is currently suited to problems that have smaller dimensionality which is due to needing to find the minimal subset of latent features that contain discriminative information and this involves examining a different combination of features. PCA, a linear dimensionality reduction approach, is used to obtain the latent features of the data which is effective for representing low dimensional data. But, this may not be effective for larger dimensional datasets where linear dimensionality reduction techniques are insufficient to find latent features that represent the data.

\section{Related work}
Counterfactual explanations are used heavily in the interpretability literature. They are useful as post-hoc local explainers which give a contrastive explanation of what needs to change to obtain a desired outcome.

One of the earliest works in counterfactual explanations was the work by Wachter et Al who proposed to perturb data points in the input space to cause it to change class, whilst using additional $L_{1}$ regularisation.
Following this, there have been many extensions that can be used to enable the generation of counterfactuals with particular properties of interest for several different purposes. One property of interest is to generate counterfactuals that lie close to the data manifold. To encourage this, there has been a significant amount of work on looking at optimizing the counterfactual in the latent space. One of the preliminary works in this area is the work by Balasubramanian et Al \cite{balasubramanian2020latent}, where they optimize for the latent features directly by passing it through an autoencoder and then examining the classification loss on the autoencoder output.
Other approaches, extend upon this idea. One example is the REVISE method by Joshi et Al which adds an additional regularization loss to encourage sparse changes \cite{joshi2019towards}. Whilst these approaches and our work both encourage the input data point to be close to the data manifold, our work perturbs the input space such that the latent features have a high likelihood, whilst the other approaches directly perturb the latent space to change its classification which can be more computationally efficient. Although, their approaches require training a generative model such as an autoencoder.

Another property desired for the counterfactuals is that they are realistic. To achieve this, there have been several approaches that look at generating counterfactuals with low uncertainty. 
One approach is the work by Schut et Al which looks at generating counterfactuals which are predicted to have a low entropy by an ensemble of classifiers \cite{schut2021generating}. In addition, the work done by Antor{\'a}n et Al calculates the uncertainty of the latent features of a data point using a Bayesian Neural Network and looks at changing the latent features such that the data point becomes more similar to a target ID class whilst also lowering the uncertainty \cite{antoran2020getting}.
These works are similar to ours as they focus on the intersection of explainability as well as uncertainty estimation. However, the purpose of these works focuses on ensuring that a data point will change class without becoming OOD whilst our work focuses on moving an OOD data point to become ID. On a practical level, these works differ from our approach as they look at making perturbation to the input in a single step, whilst our approach looks at making it change in two steps. Additionally, these approaches use an ensemble of classifiers or a Bayesian Neural Network to measure uncertainty whilst our work uses a GMM.

Another work that ensures that the counterfactual has low uncertainty is the work done by Poyiadzi et Al who introduced the FACE approach \cite{poyiadzi2020face}. The purpose of FACE is to provide users with recourse which is possible whilst there also being a path to achieve the change. This is achieved by 
generating counterfactuals in high-density regions of data space, whilst traversing a high-density path during its optimization trajectory. Both approaches constrain the path in generating counterfactuals, but, it is done in different ways. FACE ensures 
traversing along a high-density path whilst our work performs the generation in two distinct steps where we aim to have a high likelihood for both partitions at the end of the two steps.

The approach most similar to our work is the DeDUCE approach by Holtgen et Al \cite{holtgen2021deduce}. The purpose of DeDUCE is to generate counterfactuals in a computationally efficient manner. DeDUCE looks at generating counterfactuals with minimal epistemic uncertainty by perturbing data to maximize the log-likelihood where the latent feature space is modelled using a GMM. Both our approach and their approach uses a GMM to make it so that the counterfactual has a high density. However, our approach uses two different GMM (one for each type of partition) whilst they use a single partition.


\section{Conclusion}
In this work, we focused on the problem of explaining why data points are classified as OOD. This was done through the use of counterfactuals which used the concept of non-class discriminative  and  discriminative latent features to iteratively map OOD data to different OOD categories such as far-OOD, near-OOD and ID. This involved:
\begin{itemize}
    \item Introducing the concept of an OOD counterfactual.
    \item Proposing a framework for explaining OOD data points through the generation of an OOD counterfactual.
\end{itemize}
The general framework involved obtaining latent features of the data and then decomposing the latent features of the data into a non-discriminative partition, $\boldsymbol{z_n}$, and a discriminative partition $\boldsymbol{z_d}$. Following this, an OOD counterfactual is generated by performing a two-step process where each step in the process perturbs a data point in a way that makes one of the two types of latent features more similar to the ID data.
We examined the effectiveness of the OOD counterfactual on the higher dimensional tabular datasets. Compared to the baselines, it was seen that the OOD CF approach was:
\begin{itemize}
    \item Able to achieve lower Dis and Non-dis loss.
    \item Generate more realistic counterfactuals shown by the lower AUROC values.
\end{itemize}
From the ablations, it could be seen that:
\begin{itemize}
    \item Optimizing using two distinct steps leads to lower $L_{1}$ values and more realistic counterfactuals.
    \item Optimizing both steps compared to a single step is more beneficial for the lower dimensional tabular datasets.
\end{itemize}

Future work will involve working on the first two parts of the general framework. It would be desirable to develop an approach of representation learning which is both able to capture the variance in the data as well as partition the data into two distinct categories in an end-to-end manner. Additional work could also focus on partitioning the latent features into more than two categories. We hope our work will encourage others to focus on the problem of the explanation of OOD data points through the perspective of non-discriminative and discriminative latent features.

\bibliographystyle{ACM-Reference-Format}
\bibliography{ref}

\newpage
\appendix
\section{Feature partitioning algorithms}
\label{Appendix: Feature partitioning}
The overall procedure to obtain the latent feature partition consisted of many steps. Firstly, for each cardinality, we obtained the set of latent features partitions with the lowest value of $\mathcal{L}_{Partition}$ using Eqn. \ref{Eqn:Partition_loss}. This set of different latent feature partitions is denoted as ${z_{car-min}}$ and has associated loss values ${L}_{Partition-car-min}$. We obtain the mean and standard deviation of ${L}_{Partition-car-min}$ to normalize it to obtain new values denoted as ${L}_{Normalized-Partition-car-min}$. We then obtained the lowest value of ${L}_{Normalized-Partition-car-min}$ which we refer to as ${L}_{Normalized-min}$ and look choose the smallest cardinality latent feature set with a ${L}_{Normalized-Partition-car-min}$ within 10\% of  ${L}_{Normalized-min}$ as our $z_{d}$.
\section{Counterfactual generation algorithms}
\label{Appendix: Counterfactual_generation}
\begin{algorithm}[h!]
\caption{Counterfactual generation - Natural language}
\begin{algorithmic}
\State \textbf{Input:} Test sample $\boldsymbol{x}$, target class $t$, training data $\boldsymbol{X_{train}}$
\State 
\textbf{Output}: counterfactual $\boldsymbol{x'}$
\State Apply PCA on $\boldsymbol{X_{train}}$ to obtain latent features $\boldsymbol{Z_{train}}$
\State Separate $\boldsymbol{Z_{train}}$ to obtain non-discriminative 
 latent features $\boldsymbol{Z_{n,train}}$ and discriminative latent features $\boldsymbol{Z_{d,train}}$.
\State Obtain the discriminative features for the target class t, $ \boldsymbol{Z^{t}_{d, train}}$
\State Fit a GMM to $\boldsymbol{Z_{n,train}}$ and $\boldsymbol{Z^{t}_{d,train}}$
\State Decide the order of feature partitions to optimize in the two-step process
\State $x' \gets x$
\For{each partition i $\in$ Range(2)}
\State $k \gets 0$
\While{$k <$ max iter}
\State Compute the loss, $\mathcal{L}_{CF} = -\log p_{i}(f_{z}(\boldsymbol{x}))$ 
\State compute gradient g in input space, $\boldsymbol{g} = \nabla_{\boldsymbol{x}}(\mathcal{L}_{CF})$
\State $\boldsymbol{x'}= \boldsymbol{x'} - (\alpha \cdot \boldsymbol{g} $)
\State $k \gets k+1$
\EndWhile
\EndFor
\State \textbf{return} $\boldsymbol{x'}$

\end{algorithmic}
\label{alg:counterfactual_OOD_natural_language}
\end{algorithm}

\begin{algorithm}[h!]
\caption{Counterfactual generation - Pseudocode}
\begin{algorithmic}[1]

\State \textbf{Input:} Test sample $x$, target class $t$, training data $X\_train$, training labels $Y\_train$,
discriminate dimensions $dis\_dims$, non-discriminate dimensions $non\_dis\_dims$, partition order, $partitions$
\State \textbf{Output}: $counterfactual$
\State $Z\_train$ = PCA($X\_train$)
\State $Z\_n\_train$ = $Z\_train$[:,$non\_dis\_dims$]

\State  $Z\_d\_train$ = $Z\_train$[:,$dis\_dims$]
\State $class\_means$ = mean($Z\_d\_train$[:,$dis\_dims$][$Y\_train$ == c], dim=0) for c in unique($Y\_train$)])
\State $class\_covariance$ = covariance($Z\_d\_train$[:,$dis\_dims$][$Y\_train$ == c], dim=0) for c in unique($Y\_train$)])
\State $dis\_gmm$ = normal($class\_means$,$class\_covariance$) \Comment{Fit normal distribution using mean and covariance} 
\State $non\_dis\_mean$ = mean($Z_{n,train}$[:,$dis\_dims$]
\State $non\_dis\_covariance$ = covariance($Z\_n\_train$)
\State $non\_dis\_gmm$ = normal($non\_dis\_means$,$non\_dis\_covariance$) \Comment{Fit normal distribution using mean and covariance}
\State $gmms$ = order($non\_dis\_gmm$,$dis\_gmm$,$partitions$) \Comment{Order the partitions which is used in the first and second step}
\State $x' \gets x$
\For{$gmm$  $\in$ $gmms$)}
\State $k = 0$
\While{$k <$ max iter}
\State $loss$  = NLL(PCA($x$))) \Comment{Calculate negative log likelihood}
\State $g$ = backpropagate($loss$) \Comment{Backpropagate to calculate gradient}
\State $counterfactual= counterfactual - (\alpha \cdot g$)
\State $k = k+1$
\EndWhile
\EndFor
\State \textbf{return} $counterfactual$

\end{algorithmic}
\label{alg:counterfactual_OOD_pseudocode}
\end{algorithm}
\newpage

\section{Datasets}
\label{Appendix: Datasets}
\textbf{Thyroid}: The problem is to determine whether a patient referred to the clinic is hypothyroid. Therefore three classes are built: normal (not hypothyroid), hyperfunction and subnormal functioning \citep{quinlan1987inductive}.
Furthermore, we defined the subnormal functioning class as the outlier class and the other two classes are inliers, because subnormal functioning has the lowest number of data points present in the dataset.

\textbf{Diabetes}: The datasets consist of several medical predictor (independent) variables and one target (dependent) variable, Outcome. Independent variables include the number of pregnancies the patient has had, their BMI, insulin level, age, and so on. The outcomes are diabetic or non-diabetic \citep{liu2008isolation}. To define OOD data in this case, we make it so that all data points with the age attribute above the upper quartile are OOD.

\textbf{Heart}: A dataset where each row is composed of 14 attributes. 
The outcome, in this case, is whether an individual has heart disease or not \citep{kibler1989instance}. To define OOD data in this case, we make it so that all data points with the exercise-induced angina variable set to 1 are OOD.

\textbf{Wine}: These data are the results of a chemical analysis of wines grown in the same region in Italy but derived from three different cultivars. The analysis determined the quantities of 13 constituents found in each of the three types of wines \citep{kilicer2005wine}. To define OOD data in this case, we make it so that all data points classified as class 2 are OOD.

\textbf{Cardiotocography}: 2126 fetal cardiotocograms (CTGs) were automatically processed and the respective diagnostic features were measured. The CTGs were also classified by three expert obstetricians and a consensus classification label was assigned to each of them. Classification was with respect to a fetal state (N, S, P) \citep{ayres2000sisporto}. To define OOD data in this case, we make it so that all data points classified as P are OOD.

\end{document}